\documentclass[10pt,twocolumn,letterpaper]{article}

\usepackage{cvpr}              

\usepackage{graphicx}
\usepackage{amsmath}
\usepackage{amssymb}
\usepackage{booktabs}
\usepackage{algorithm}
\usepackage{algpseudocode} 
\usepackage{amsmath}
\usepackage{multirow}
\usepackage{cases}
\usepackage{booktabs}
\usepackage{color}
\usepackage{cases}
\usepackage{wasysym}
\usepackage[colorlinks,linkcolor=blue]{hyperref}
\usepackage[accsupp]{axessibility}

\usepackage[capitalize]{cleveref}
\crefname{section}{Sec.}{Secs.}
\Crefname{section}{Section}{Sections}
\Crefname{table}{Table}{Tables}
\crefname{table}{Tab.}{Tabs.}

\def\cvprPaperID{*****} 
\def\confName{CVPR}
\def\confYear{2023}

\begin{document}

\title{MMANet: Margin-aware Distillation and Modality-aware Regularization for Incomplete Multimodal Learning}  

\author{Shicai Wei \quad\quad\quad  Chunbo Luo \quad\quad\quad  Yang Luo  \\
School of Information and Communication Engineering \\
University of Electronic Science and Technology of China\\
{\tt\small shicaiwei@std.uestc.edu.cn  \{c.luo, luoyang\}@uestc.edu.cn }
}
\maketitle

\begin{abstract}
Multimodal learning has shown great potentials in numerous scenes and attracts increasing interest recently. However, it often encounters the problem of missing modality data and thus suffers severe performance degradation in practice. To this end, we propose a general framework called MMANet to assist incomplete multimodal learning. It consists of three components: the deployment network used for inference, the teacher network transferring comprehensive multimodal information to the deployment network, and the regularization network guiding the deployment network to balance weak modality combinations. Specifically, we propose a novel margin-aware distillation (MAD) to assist the information transfer by weighing the sample contribution with the classification uncertainty. This encourages the deployment network to focus on the samples near decision boundaries and acquire the refined inter-class margin. Besides, we design a modality-aware regularization (MAR) algorithm to mine the weak modality combinations and guide the regularization network to calculate prediction loss for them. This forces the deployment network to improve its representation ability for the weak modality combinations adaptively. Finally, extensive experiments on multimodal classification and segmentation tasks demonstrate that our MMANet outperforms the state-of-the-art significantly. Code is available at: \href{https://github.com/shicaiwei123/MMANet}{https://github.com/shicaiwei123/MMANet}
\end{abstract}

\section{Introduction}
\label{Introduction}

Multimodal learning has achieved great success on many vision tasks such as classification~\cite{mm_cf1,mm_cf2,mm_cf3}, object detection~\cite{mm_detection1,mm_detection2,mm_detection3}, and segmentation~\cite{rgbd_seg1,rgbd_seg2,rgbd_seg3}. However, most successful methods assume that the models are trained and tested with the same modality data. In fact, limited by device~\cite{spcical-gan1,device1}, user privacy~\cite{privacy1,privacy2}, and working condition~\cite{special-hall1,special-hall2}, it is often very costly or even infeasible to collect complete modality data during the inference stage. There is thus substantial interest in assisting the incomplete or even single modality inference via the complete modality data during training.

A typical solution is to reconstruct the sample or feature of the missing modalities from the available ones~\cite{special-hall2,special-hall3,special-hall4,special-hall5,special-hall7,spcical-gan1}. Nevertheless, this needs to build a specific model for each modality from all possible modality combinations and thus has high complexity. Recent studies focus on learning a unified model, instead of a bunch of networks, for different modality combinations. Generally, many such approaches~\cite{rfnet,mmformer,robust,hemis,hetero,lcr} attempt to leverage feature fusion strategies to capture modality-invariant representation so that the model can adapt to all possible modality combinations. For example, RFNet~\cite{rfnet} designs the region-aware fusion module to fuse the features of available image modalities.

\begin{table}[]
\centering
\renewcommand\tabcolsep{8.0pt}
\begin{tabular}{cccc}
\toprule[1pt]
Modality   & Customized & Unified & Drop \\ \toprule[1pt]
RGB       & 10.01 & 11.75     & -1.65 \\ 
Depth      & 4.45 & 5.87      & -1.42  \\ 
IR        &11.65 & 16.62    &  -4.97  \\ 
RGB+Depth    & 3.41 & 4.61     &  -1.2  \\ 
RGB+IR      & 6.32  & 6.68     & -0.36  \\ 
Depth+IR     & 3.54  & 4.95     & -1.41  \\ 
RGB+Depth+IR    & 1.23 & 2.21     & -0.98  \\ \toprule[1pt]
\end{tabular}
\caption{The performance of customized models and the unified model for different modality combinations on the CASIA-SURF dataset using the average classification error rate. The `customized` means to train a model for each combination independently while the `unified' means to train only one model for all the combinations. The architectures of all the models are the same and the feature map of missing modality (such as the IR for RGB+Depth) is set as zero.}
\label{PVU}
\end{table}

Although the existing unified models are indeed able to increase the efficiency of training and deployment of the multimodal models, their performance is likely to be sub-optimal. As shown in Table~\ref{PVU}, the customized models consistently outperform the unified model for different modality combinations. This is because existing unified models usually focus on the modality-invariant features while ignoring the modality-specific information. Note that the complementary modality-specific information of multiple modalities can help refine the inter-class discrimination and improve inference performance~\cite{ubm1,ubm2,multimodal-survey}. This motivates us to propose the first research question of this paper: \textbf{Can a unified model consider the modality invariant and specific information simultaneously while maintaining robustness for incomplete modality input? }

To this end, we propose to guide the unified model to learn the comprehensive multimodal information from the teacher model trained with complete modality. This regularizes the target task loss to encourage the unified model to acquire complementary information among different modality combinations multimodal information while preserving the generalization to them. Specifically, we propose a novel margin-aware distillation (MAD) that trains the unified model by guiding it to mimic the inter-sample relation of the teacher model. MAD introduces the classification uncertainty of samples to re-weigh their contribution to the final loss. Since the samples near the class boundary are more likely to be misclassified and have higher classification uncertainty~\cite{margin1}, this encourages the unified model to preserve the inter-class margin refined by the complementary cues and learn the modality-specific information.


Another limitation of existing unified approaches is that they struggle to obtain optimal performance for the unbalanced training problem. To be specific, conventional multimodal learning models tend to fit the discriminative modality combination and their performance will degrade significantly when facing weak modality combinations. To solve this issue, existing unified approaches introduce the auxiliary discriminator to enhance the discrimination ability of the unimodal combinations~\cite{rfnet,mmformer,robust}. This utilizes a hypothesis that a single modality is weaker than multiple ones. However, as shown in Table~\ref{PVU}, no matter for the customized model or the unified model, the single Depth modality outperforms the RGB, IR, and their combinations. This indicates the combination with multiple weak modalities may be harder to be optimized than a single strong modality. Moreover, as shown in Table~\ref{pe-c-cefa}, RGB becomes the strong modality while Depth and IR become the weak modalities. This indicates that the modality importance is not fixed but varies with scenarios. These findings motivate us to propose the second research question: \textbf{How to effectively optimize the weak modality combination in varying scenarios?}

To this end, we design a regularization network and MAR algorithm to assist the training of the unified network. Specifically, the regularization network generates additional predictions for all inputs. Then MAR mines and calculates prediction loss for the sample from the weak combinations. This forces the unified model to improve its representation ability for the weak combination. In detail, MAR mines the weak combination via the memorization effect~\cite{mem1,mem2,mem3} that DNNs tend to first memorize simple examples before overfitting hard examples. As shown in Fig.~\ref{training_pro}(a), the unified model tends to fit the samples containing Depth modality firstly at the early stage. Therefore, MAR first mines the strong modality via the memorization effect. Then it determines the combinations of rest modalities as the weak ones.


Finally, we develop a model and task agnostic framework called MMANet to assist incomplete multimodal learning by combining the proposed MAD and MAR strategies. MMANet can guide the unified model to acquire comprehensive multimodal information and balance the performance of the strong and weak modality combination simultaneously. Extensive comparison and ablation experiments on multimodal classification and segmentation tasks demonstrate the effectiveness of the MMANet.

\section{Related work}

\subsection{Multimodal Learning for Missing Modalities}

Most existing multimodal learning methods assume that all instances consist of full modalities. However, this assumption does not always hold in real-world applications due to the device~\cite{spcical-gan1,device1}, user privacy~\cite{privacy1,privacy2}, and working condition~\cite{special-hall1,special-hall2}. Recently, many incomplete multimodal learning methods have been proposed and can be roughly categorized into two types: customized methods and unified methods. Customized methods aim to train a specific model to recover the missing modality in each incomplete modality combination. According to the recovering target, the customized methods can be further divided into sample-based methods and representation-based methods. Sample-based methods focus on imputing the missing modality at the input space with generative adversarial networks~\cite{MC2,MC3,spcical-gan1,special-gan2}. Due to the complexity of sample reconstruction, it is usually unstable and may introduce noise to harm the primary task at hand~\cite{gan-issue1}. Thus the representation-based methods are proposed to reconstruct the sample representation via the knowledge distillation~\cite{special-hall1,special-hall2,special-hall3,special-hall4} or matrix completion~\cite{lmc1,lmc2}. Although promising results are obtained, these methods have to train and deploy a specific model for each subset of missing modalities, which has high complexity in practical applications.

The unified methods aim to train one model to deal with different incomplete modality combinations by extracting the modality-invariant features. For example, HeMIS~\cite{hemis} learns an embedding of multimodal information by computing statistics (i.e., mean and variance) from any number of available modalities. Furthermore, Chen~\etal introduce the feature disentanglement to cancel out the modality-specific information. Besides, more recent methods, such as LCR~\cite{lcr} and RFNet~\cite{rfnet} focus on extracting the modality-invariant representation via different attention mechanisms. Moreover, mmFormer~\cite{mmformer} introduces the transformer block to model the global semantic information for the modality-invariant embedding. While these methods achieve promising results, they only consider the modality-invariant information while ignoring the modality-specific information. As a result, they usually perform much worse than the customized methods, especially when more than one modality is missing~\cite{acn}.



\begin{figure*}[ht]
\centering
\includegraphics[width=0.9\textwidth]{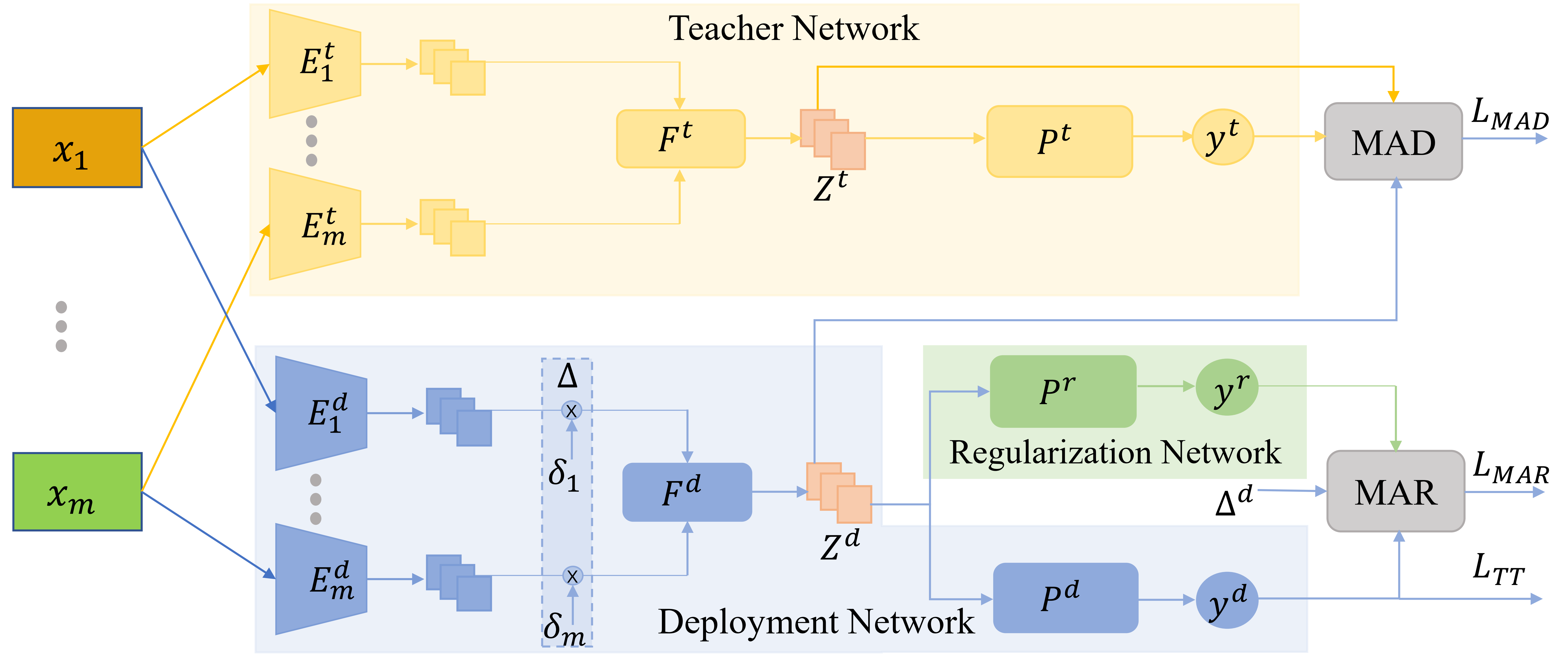} 
\caption{Overview of the proposed MMANet. It consists of three parts: the deployment network used for final inference, the teacher network transferring comprehensive multimodal knowledge to the deployment network, and the regularization network guiding the deployment network to balance weak modality combinations.}
\label{MMANET}
\end{figure*}


\subsection{Knowledge Distillation}

Knowledge distillation aims to transfer knowledge from a strong teacher to a weaker student network to facilitate supervised learning. Generally, the distillation method can be divided into three types: response-based distillation that matches the softened logits of teachers and students~\cite{Distilling}, the representation-based distillation that matches the feature maps~\cite{hint,hint1,hint2}, and the relation-based distillation that matches the sample relations.~\cite{pkt,relation-sp}.


While originating from the resource-efficient deep learning, knowledge distillation has found wider applications in such areas as incomplete multimodal learning. Here, it is used to transfer the privileged modality information that can only be accessed during the training stage from the teacher to the student~\cite{special-hall1,special-hall2}. Since the input of the teacher and student network is different in incomplete multimodal learning, transferring knowledge by representation-based methods may lead to overfitting~\cite{special-hall7}. Recent methods focus on transferring the privileged modality information by the relation-based methods~\cite{acn,acn2,acn3}. However, these prior arts usually consider different instances equally and ignore their specificity, which would lead to sub-optimal performance.

\section{Method}

\subsection{MMANet}

In this section, we introduce a general framework called MMANet to address the challenge of incomplete multimodal learning. As shown in Fig.~\ref{MMANET}, it consists of three parts: deployment network, teacher network, and regularization network. Specifically, the deployment network is the inference network. To make it robust to the modality incompleteness, MMANet introduces the Bernoulli indicator $\Delta=\{\delta_{1}...\delta_{m}\}$ after modality encoders and conducts modality dropout during the training stage by randomly setting some components of $\Delta$ as 0. For missing modalities, the corresponding encoded feature maps will be replaced by a zero matrix. Besides, MMANet introduces the teacher network that is pre-trained with complete multimodal data to transfer the comprehensive multimodal knowledge to the deployment network via the MAD. This helps the deployment network acquire the modality-invariant and specific features simultaneously. Finally, MMANet guides the deployment network to train together with the regularization network that produces additional predictions for the weak modality combination via the MAR. This alleviates the overfitting for strong modality combinations. The total loss to guide the training of the deployment network is defined as follows,


\begin{equation}
  L_{total}=L_{TL}+ \alpha L_{MAD}+\beta L_{MAR}
\end{equation} where $\alpha$ and $\beta$ are the hyper-parameters. $L_{LT}$ is task learning loss, which is determined by the primary task at hand. For example, $L_{LT}$ may be the cross entropy loss when the primary task is classification. $L_{MAD}$ and $L_{MAR}$ are the loss of MAD and MAR respectively. 

Besides, the other nations used in MMANet are defined as follows. Given a mini-batch multimodal input $x=\{x_{1},...,x_{m}\}$, $x_{m} \in R^{b}$ denotes the data of $m_{th}$ modality. $b$ is the batch size. $E_{m}^{t}$ and $E_{m}^{d}$ denote the encoders for the $m_{th}$ modality in the teacher and deployment networks, respectively. $F^{t}$ and $F^{d}$ denote the fusion module used in the teacher and deployment networks, respectively. $\Delta^{d}\in R^{b\times m}$ is the vector of $\Delta$. $z^{t} \in R^{b^{t} \times c^{t} \times h^{t} \times w^{t}}$ and $z^{d} \in R^{b^{d} \times c^{d} \times h^{d} \times w^{d}}$ denote the fused features of the teacher and deployment networks, respectively. Here, where $b$ is the batch size, $c$ is the number of output channels, and $h$ and $w$ are spatial dimensions. $P^{t}$, $P^{r}$, and $P^{d}$ denote the task predictor of the teacher, regularization, and deployment networks, respectively. $y^{t}$,$y^{r}$, and $y^{d}$ denote the $R^{b \times k}$ prediction matrix of the teacher, regularization, and deployment networks, respectively. Here, $k$ is the class number.

\begin{figure}[t]
\centering
\includegraphics[width=1.0\columnwidth]{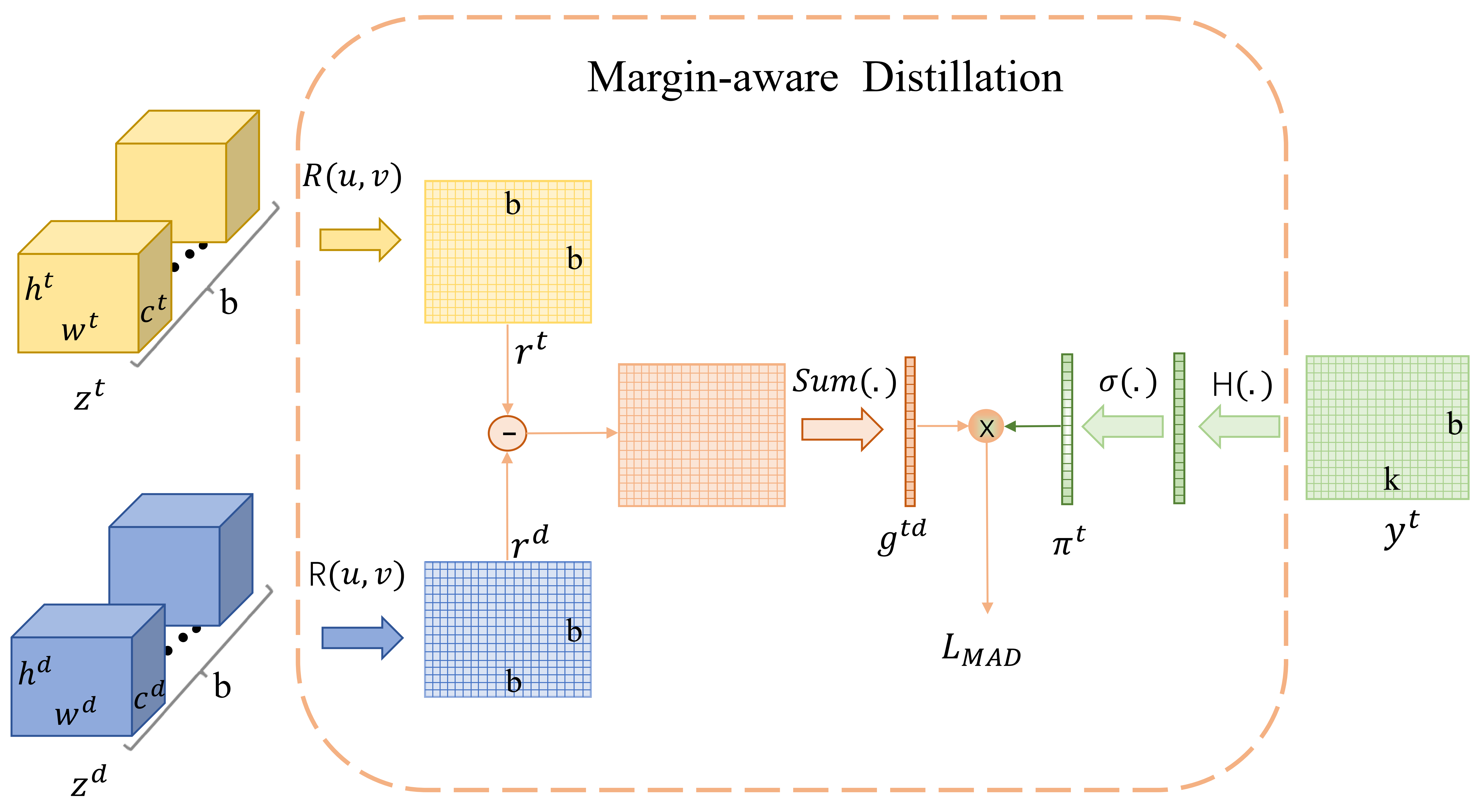} 
\caption{The illustration of the proposed MAD.}
\label{MAD}
\end{figure}

\subsection{MAD}

This section introduces the proposed MAD strategy for transferring the comprehensive multimodal information from the teacher network to the deployment network. As shown in Fig.~\ref{MMANET}, MAD is conducted between the $z^{t}$ and $z^{d}$. $z^{d}$ of a simple is varying due to the random modality dropout. In contrast, the sample semantic is invariant. Thus, MAD proposes to transfer the teacher's knowledge via relation consistency instead of feature consistency. This helps avoid overfitting and harming the representation ability of deployment networks. Moreover, MAD proposes to measure the class boundaries and guide the unified model to pay more attention to the samples near them. This can encourage the development network to inherit the refined inter-class margin from the teacher network. Nevertheless, boundaries are usually difficult to detect due to their irregularity. To solve this issue, MAD introduces the classification uncertainty of each sample to re-weight its contribution for the total loss. Since the samples near the class boundaries are more likely to be misclassified and have higher classification uncertainty, this can realize attention to them.

The overview of the MAD is shown in Fig.~\ref{MAD}. It takes $z^{t}$, $z^{d}$, and $y^{t}$ as the input and consists of three steps: (a) calculating the relation discrepancy vector $g^{td} \in R^{b}$, (b) calculating the classification uncertainty vector $\pi^{t} \in R^{b}$ and (c) calculating the total loss $ L_{MAD}$ for MAD. 

(a) MAD calculates $g^{td}$ from $z^{t}$ and $z^{d}$. Specifically, MAD first reshape $z^{t}$ and $z^{d}$ into $z^{t^{\prime}} \in R^{b^{t} \times c^{t} * h^{t} * w^{t}}$ and $z^{d^{\prime}} \in R^{b^{d} \times c^{d} * h^{d} * w^{d}}$. Then MAD calculates the relation matrix $r^{t} \in R^{b \times b}$ and $r^{d} \in R^{b \times b}$ via the same relation function R(u,v), respectively. And the ${r}^{t}(i,j)$ that denotes the relation between $i_{th}$ and ${j_{th}}$ sample representations of the teacher network can be expressed as follows,


\begin{equation}
  {r}^{t}(i,j)=R(z^{t^{\prime}}(i,:),z^{t^{\prime}}(j,:))
\end{equation}

Besides, the ${r}^{d}(i,j)$ that denotes the relation between $i_{th}$ and ${j_{th}}$ sample representations of the deployment network can be expressed as follows,

\begin{equation}
r^{d}(i,j)=R(z^{d^{\prime}}(i,:),z^{d^{\prime}}(j,:))
\end{equation}

Theoretically, $R(u,v)$ can be any metric for measuring the vector distance, such as the Euclidean distance and the cosine distance. Because the dimension of the feature vectors of the teacher and the deployment networks could be very high, to eliminate the curse of dimensionality, we choose cosine distance as the $R(u,v)$,

\begin{equation}
  R(u,v)=\frac{{u}^{T} {v}}{\|{u}\|_{2}|| {v} \|_{2}}
\end{equation}

Furthermore, MAD calculates the discrepancy matrix between $r^{t}$ and $r^{d}$ and sum each row to get $g^{td}$.

\begin{equation}
g^{td}= \sum_{i=1}^{b} (r^t-r^d)_{i}
\end{equation}

Here, $g^{td}(i)$ denotes the relation gap between the teacher and deployment networks from the $i_{th}$ sample to other samples in the same mini-batch. 

(b) MAD calculates $\pi^{t}$ from $y^t$. In detail, it takes the information entropy of the logit output of each sample as its classification uncertainty. And the classification uncertainty for $i_{th}$ sample, $\pi^{t}(i)$, can be expressed as follows,

\begin{numcases}{}
  \pi^{t}(i)=H(y^{t}(i,:))\\
  H(x)=-\sigma(x)*log(\sigma(x))
\end{numcases} where $\sigma(.)$ is the softmax function for normalization. $H(x)$ is the information entropy of $x$. A sample that has a higher classification uncertainty is usually closer to the decision boundaries, since it is more likely to be misclassified. Thus, $\pi^{t}(i), i \in [1,b]$ can also denote the margin from the $i_{th}$ sample representation to the decision boundary.

(c) Finally, MAD takes $\pi^{t}(i)$ as the weight for the corresponding component $g^{td}(i)$ to calculate $ L_{MAD}$,

\begin{equation}
  L_{MAD}=\sum_{i=1}^{b} \sigma(\pi^{t})(i)* g^{td}(i)
\end{equation} This encourages the deployment network to focus on the samples near the decision boundaries and preserve the inter-class margin refined by the comprehensive multimodal information from the teacher network.

\subsection{MAR}


This section introduces the MAR algorithm that forces the deployment network to improve its discriminating ability for weak modality combinations adaptively. As shown in Fig.~\ref{MMANET}, MAR takes the $y^{r}$, $y^{d}$ and $\Delta^{d}$ as as the input to calculate the $L_{MAR}$. Specifically, MAR first proposes a contrastive ranking strategy to mine the weak modality combinations. Compared to simply taking the combination with a single modality as the weak one, this further considers the combination with multiple modalities and can get more accurate mining results. Then, MAR calculates the prediction loss for the weak modality combinations, guiding the deployment network to pay more attention to them.

\begin{figure}[t]
\centering
\includegraphics[width=1.0\columnwidth]{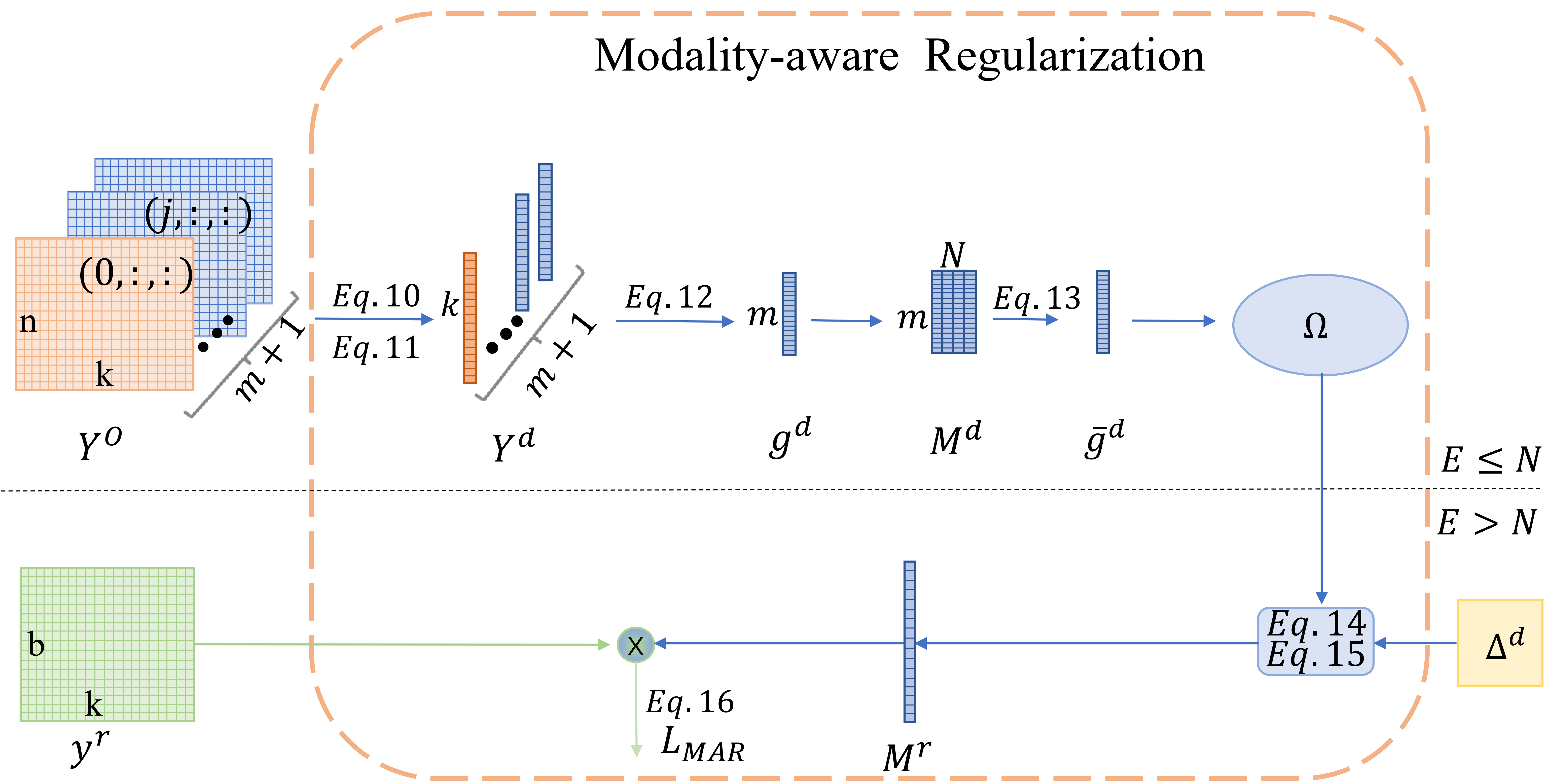} 
\caption{The illustration of the proposed MAR.}
\label{MAR}
\end{figure}

The overview of MAR is shown in Fig.~\ref{MAR}. It consists of two steps: (a) when $E\leq N$, mining the weak modality combination set $\Omega$, and (b) when $E > N$, calculating $L_{MAR}$. Here $E$ is the current training epoch, and $N$ is the number of warm-up epochs.




%

(a) MAR calculates $\Omega$ from $y^{d}$ using contrastive ranking. MAR proposes to calculate the predicted output $Y^{O} \in R^{(m+1)\times n \times k}$ of $\Delta_{i}$,  $i \in [0,m]$, on the train set after each training epoch.
\begin{equation}
  Y^{O}(i,:,:)=y^{d}(\Delta_{i})
\end{equation} $n$ is sample number. $\Delta_{i}$ means the $i_{th}$ component of $\Delta$ is 0. $\Delta_{0}$ means none component of $\Delta$ are 0, which must contain the strong modality. Since the deployment network tends to first memorize the samples with strong modality, $\Delta_{w},w\in[1,m]$ that makes $Y^{O}(w,:,:)$ has the biggest distance with $Y^{O}(0,:,:)$ is the hard combination that does not contain the strong modality. And the element of $\Omega$ can be determined as $\Delta_{w}$ and the $\Delta$ consists of the modalities in it.


Specifically, to make $\Delta_{w}$ robust for the randomness of neural network learning, MAR introduces two innovations. Firstly, MAR calculates the prediction discrepancy from the prediction distribution $Y^{d} \in R^{(m+1)\times k}$ instead of $Y^{O}$,
\setlength\abovedisplayskip{0.3cm}
\setlength\belowdisplayskip{0.3cm}
\begin{numcases}{}
Y^{d}(i,j)=\sum (Y^{D}(i,:)==j)\\
Y^{D}=\mathop{\arg\max}(Y^{O},dim=2)
\end{numcases} where $j\in[0, k-1]$. Compared with $Y^{O}$, $Y^{d}$ needs only class-wise but not sample-wise consistency. Then the vector discrepancy vector $g^{d} \in R^{m}$ is defined as follows,







\setlength\abovedisplayskip{0.1cm}
\setlength\belowdisplayskip{0.3cm}
\begin{equation}
\label{gd}
  g^{d}(i)=KL(log(\sigma({Y}^{d}(i))),\sigma({Y}^{d}(0)))
\end{equation} where KL(,) means the KL divergence, $i \in [1,m]$. 

Secondly, MAR introduces a memory bank $M^{d} \in R^{N \times m}$ to save the $g^{d}$ among the warm-up epochs and performs average filtering to obtain $\overline{g}^{d}$,

\setlength\abovedisplayskip{0.0cm}
\setlength\belowdisplayskip{0.2cm}
\begin{equation}
  \overline{g}^{d} = \sum_{i=1}^{N} \frac{1}{N} (M^{d})_{i}
\end{equation} where $(M^{d})_{i}$ is the $g^{d}$ in the $i_{th}$ epoch. And $\Delta_{w}$ can be determined as $\Delta_{i}$ where $i=argmax(\overline{g}^{d})$.



(b) MAR calculates $L_{MAR}$ from $y^{r}$, $\Delta^{d}$ and $\Omega$. In detail, MAR first calculates the weak combination mask $M^{r} \in R^{b}$ from $\Delta^{d}$ and $\Omega$,
\setlength\abovedisplayskip{0.1cm}
\setlength\belowdisplayskip{0.3cm}
\begin{numcases}{}
M^{r}(i)=FALSE \quad if \Delta^{d}(i) \notin \Omega \\
M^{r}(i)=TRUE \quad if \Delta^{d}(i) \in \Omega
\end{numcases} where $i \in [0,b-1]$, $\Delta^{d}(i)$ is the $\Delta$ for the $i_{th}$ sample in this mini-batch. Then, the $L_{MAR}$ is defined as follows,

\setlength\abovedisplayskip{0.1cm}
\setlength\belowdisplayskip{0.3cm}
\begin{equation}
  L_{MAR}=L_{TL}(y^{r}[M^{r}],l[M^{r}])
\end{equation} where $l$ is the groundtruth vector for $y^{r}$. $[.]$ denotes the index operator.

\section{Experiments}

We conduct experiments on multimodal classification and segmentation tasks to evaluate the proposed MMANet. In the following, we first compare the MMANet architecture with the state-of-the-art on these two tasks. Then, we ablate the MAD and MAR strategies of MMANet.

\begin{table*}[h]

\caption{Performance on the multimodal classification task with CASIA-SURF. $\downarrow$ means that the lower the value, the better the performance.}
\label{pe-c-surf}

\centering\begin{tabular}{ccc|ccccccc}
\toprule
\multicolumn{3}{c|}{Modalities}                                         & \multicolumn{7}{c}{ACER($\downarrow$)}                                                                                             \\ \hline
\multicolumn{1}{c}{}           & \multicolumn{1}{c}{}           &           & \multicolumn{1}{c|}{Customized} & \multicolumn{6}{c}{Unified}                                                                           \\ \cline{4-10} 
\multicolumn{1}{c}{\multirow{-2}{*}{RGB}} & \multicolumn{1}{c}{\multirow{-2}{*}{Depth}} & \multirow{-2}{*}{IR} & \multicolumn{1}{c|}{SF}    & \multicolumn{1}{c}{SF-MD} & \multicolumn{1}{c}{HeMIS}            & \multicolumn{1}{c}{LCR}  & \multicolumn{1}{c}{RFNet} & \multicolumn{1}{c|}{MMFormer} & MMANet \\ \toprule
\multicolumn{1}{c}{\CIRCLE}           & \multicolumn{1}{c}{\Circle}           &     \multicolumn{1}{c|}{\Circle}        & \multicolumn{1}{c|}{10.01}   & \multicolumn{1}{c}{11.75} & \multicolumn{1}{c}{{14.36}} & \multicolumn{1}{c}{13.44} & \multicolumn{1}{c}{12.43} & \multicolumn{1}{c|}{11.15}  & \textbf{8.57}  \\ 
\multicolumn{1}{c}{\Circle}           & \multicolumn{1}{c}{\CIRCLE}           &       \multicolumn{1}{c|}{\Circle}      & \multicolumn{1}{c|}{4.45}   & \multicolumn{1}{c}{5.87} & \multicolumn{1}{c}{4.70}             & \multicolumn{1}{c}{4.40} & \multicolumn{1}{c}{4.17} & \multicolumn{1}{c|}{4.67}   & \textbf{2.27}  \\ 
\multicolumn{1}{c}{\Circle}           & \multicolumn{1}{c}{\Circle}           & \CIRCLE          & \multicolumn{1}{c|}{11.65}   & \multicolumn{1}{c}{16.62} & \multicolumn{1}{c}{16.21}            & \multicolumn{1}{c}{15.26} & \multicolumn{1}{c}{14.69} & \multicolumn{1}{c|}{13.99}  & \textbf{10.04} \\ 
\multicolumn{1}{c}{\CIRCLE}           & \multicolumn{1}{c}{\CIRCLE}           &       \multicolumn{1}{c|}{\Circle}      & \multicolumn{1}{c|}{3.41}   & \multicolumn{1}{c}{4.61} & \multicolumn{1}{c}{3.23}             & \multicolumn{1}{c}{3.32} & \multicolumn{1}{c}{2.23} & \multicolumn{1}{c|}{1.93}   & \textbf{1.61}  \\ 
\multicolumn{1}{c}{\CIRCLE}           & \multicolumn{1}{c}{\Circle}           & \CIRCLE          & \multicolumn{1}{c|}{6.32}   & \multicolumn{1}{c}{6.68} & \multicolumn{1}{c}{6.27}             & \multicolumn{1}{c}{5.16} & \multicolumn{1}{c}{4.27} & \multicolumn{1}{c|}{4.77}   & \textbf{3.01}  \\ 
\multicolumn{1}{c}{\Circle}           & \multicolumn{1}{c}{\CIRCLE}          & \CIRCLE         & \multicolumn{1}{c|}{3.54}   & \multicolumn{1}{c}{4.95} & \multicolumn{1}{c}{3.68}             & \multicolumn{1}{c}{3.53} & \multicolumn{1}{c}{3.22} & \multicolumn{1}{c|}{3.10}   & \textbf{1.18}  \\ 
\multicolumn{1}{c}{\CIRCLE}           & \multicolumn{1}{c}{\CIRCLE}           & \CIRCLE          & \multicolumn{1}{c|}{1.23}   & \multicolumn{1}{c}{2.21} & \multicolumn{1}{c}{1.97}             & \multicolumn{1}{c}{1.88} & \multicolumn{1}{c}{ 1.18} & \multicolumn{1}{c|}{1.94}   & \textbf{ 0.87}  \\ \toprule
\multicolumn{3}{c|}{Average}                                           & \multicolumn{1}{c|}{5.80}   & \multicolumn{1}{c}{7.52} & \multicolumn{1}{c}{7.18}             & \multicolumn{1}{c}{6.71} & \multicolumn{1}{c}{6.02} & \multicolumn{1}{c|}{5.93}   & \textbf{3.94}  \\ \toprule
\end{tabular}

\end{table*}

\begin{table}[]
\setlength\tabcolsep{4pt} 
\caption{Performance on the multimodal classification task with the CeFA dataset.}
\label{pe-c-cefa}
\begin{tabular}{ccc|ccc}
\toprule
\multicolumn{3}{c|}{Modalities}                                        & \multicolumn{3}{c}{ACER($\downarrow$)}                        \\ \hline
\multicolumn{1}{c}{\multirow{2}{*}{RGB}} & \multicolumn{1}{c}{\multirow{2}{*}{Depth}} & \multirow{2}{*}{IR} & \multicolumn{1}{c|}{Customized} & \multicolumn{2}{c}{Unified}      \\ \cline{4-6} 
         &           &  \multicolumn{1}{c|}{}         & \multicolumn{1}{c|}{SF}   & \multicolumn{1}{c|}{MMFormer} & MMANet \\ \toprule
\multicolumn{1}{c}{\CIRCLE}          & \Circle            &  \Circle         & \multicolumn{1}{c|}{27.44}   & \multicolumn{1}{c|}{28.51}  & \textbf{27.15} \\ 
\multicolumn{1}{c}{\Circle}           & \multicolumn{1}{c}{\CIRCLE}           &    \Circle        & \multicolumn{1}{c|}{33.75}   & \multicolumn{1}{c|}{33.58}  & \textbf{32.50} \\ 
\multicolumn{1}{c}{\Circle }          & \Circle            & \CIRCLE         & \multicolumn{1}{c|}{36.17}   & \multicolumn{1}{c|}{39.56}  & \textbf{35.62} \\ 
\multicolumn{1}{c}{\CIRCLE}          & \multicolumn{1}{c}{\CIRCLE}           &   \Circle        & \multicolumn{1}{c|}{35.62}   & \multicolumn{1}{c|}{29.47}  & \textbf{22.87} \\ 
\multicolumn{1}{c}{\CIRCLE}          &   \Circle            & \CIRCLE         & \multicolumn{1}{c|}{31.62}   & \multicolumn{1}{c|}{27.66}  & \textbf{23.27} \\ 
\multicolumn{1}{c}{\Circle}           & \multicolumn{1}{c}{\CIRCLE}           & \CIRCLE         & \multicolumn{1}{c|}{36.62}   & \multicolumn{1}{c|}{32.17}  & \textbf{30.45} \\ 
\multicolumn{1}{c}{\CIRCLE}          & \multicolumn{1}{c}{\CIRCLE}           & \CIRCLE         & \multicolumn{1}{c|}{24.15}   & \multicolumn{1}{c|}{30.72}  & \textbf{23.68} \\ \toprule
\multicolumn{3}{c|}{Average}                                         & \multicolumn{1}{c|}{32.20}   & \multicolumn{1}{c|}{31.52}  & \textbf{27.94} \\ \toprule
\end{tabular}
\end{table}

\subsection{Performance and Comparison on Multimodal Classification}

\textbf{Datasets}: We take the face anti-spoofing task as the example of the multimodal classification and conduct experiments on the CASIA-SURF~\cite{surf} and CeFA~\cite{cefa} datasets. Both of them consist of the RGB, Depth, and IR modalities. For CASIA-SURF, we follow the intra-testing protocol suggested by the authors and divide it into train, validation, and test sets with 29k, 1k, and 57k samples, respectively. For CeFA, we follow the cross-ethnicity and cross-attack protocol suggested by the authors and divide it into train, validation, and test sets with 35k, 18k, and 54k samples, respectively. Here we report the results on the test set using the metric of Average Classification Error Rate (ACER)~\cite{surf}.


\textbf{Implementation}: We use random flipping, rotation, and cropping for data augmentation. All models are optimized by an SGD for 100 epochs with a mini-batch 64. The learning rate is initialized to 0.001 with 5 epochs of linear warm-up and divided by 10 at 16, 33, and 50 epochs. Weight decay and momentum are set to 0.0005 and 0.9, respectively.

The hyper-parameters of comparison methods use the suggested ones in the original articles. The $(\alpha,\beta)$ for MMANet is set as $(30,0.5)$ and $(30,0.5)$ for CASIA-SURF and CeFA, respectively. The warm-up epoch $N$ is set as 5.

\textbf{Comparison}: Here we compare MMANet with two different unified methods for incomplete multimodal learning. One is an early method that only focuses on extracting modality-invariant features, such as HeMIS~\cite{hemis} and LCR~\cite{lcr}. Another is the enhanced method that further considers improving the discrimination ability for single-modal combinations, such as RFNet~\cite{rfnet}, and mmFormer~\cite{mmformer}. 

Besides, we introduce two baseline methods, SF and SF-MD. SF~\cite{surf} is the benchmark method of the CASIA-SURF, which is a customized method that trains the model for each modality combination. SF-MD is the variant of SF by simply adding the Bernoulli indicator after its modality encoder. This enables SF-MD to become a unified model that trains a single model for all modality combinations.

Finally, for a fair comparison, we follow the basic implementation of SF for all the comparison methods. Specifically, we unify the modality encoders of HeMIS, LCR, RFNet, and mmFormer as the ResNet18 used in SF. Besides, we set the SF model trained with complete multimodal data as the teacher network and the SF-MD model as the development network.


\textbf{Results}: Table~\ref{pe-c-surf} and Table~\ref{pe-c-cefa} show the comparison results with the state-of-the-art methods on the CASIS-SURF and CeFA datasets, respectively. Compared with the second-best unified method, i.e. mmFormer, MMANet decreases the average ACER by 1.99\% and 3.58\% on the CASIS-SURF and CeFA, respectively. Besides, we can see that MMANet achieves the best performance on both datasets for all the nine modality combinations for CASIA-SURF. This shows the superiority of our method on the incomplete multimodal classification task. More importantly, MMANet even outperforms the customized baseline method, i.e. SF, for all the modality combinations on the CASIA-SURF and CeFA, decreasing the average ACER by 1.86\% and 4.26\%. This demonstrates the effectiveness of the proposed MAD and MAR for the incomplete multimodal classification task.



\begin{table*}[]
\centering
\caption{Performance on the multimodal segmentation task with NYUv2. $\uparrow$ means that the higher the value, the better the performance.}
\label{pe-se-nyu}
\begin{tabular}{cc|ccccccc}
\toprule
\multicolumn{2}{c|}{Modality}                   & \multicolumn{7}{c}{mIOU($\uparrow$)}                                                                                    \\ \toprule
\multicolumn{1}{c}{\multirow{2}{*}{RGB}} & \multirow{2}{*}{Depth} & \multicolumn{1}{c|}{Customized} & \multicolumn{6}{c}{Unified}                                                                  \\ \cline{3-9} 
\multicolumn{1}{c}{}           &            & \multicolumn{1}{c|}{ESANet}  & \multicolumn{1}{c}{ESANet-MD} & \multicolumn{1}{c}{HeMIS} & \multicolumn{1}{c}{LCR}  & \multicolumn{1}{c}{RFNet} & \multicolumn{1}{c|}{mmFormer} & MMANet \\ \toprule
\multicolumn{1}{c}{\CIRCLE}         &    \Circle         & \multicolumn{1}{c|}{44.22}   & \multicolumn{1}{c}{41.34}   & \multicolumn{1}{c}{33.23} & \multicolumn{1}{c}{41.91} & \multicolumn{1}{c}{42.89} & \multicolumn{1}{c|}{43.22}  & \textbf{44.93} \\ 
 \Circle             & \CIRCLE          & \multicolumn{1}{c|}{40.55}   & \multicolumn{1}{c}{39.76}   & \multicolumn{1}{c}{31.23} & \multicolumn{1}{c}{39.88} & \multicolumn{1}{c}{40.76} & \multicolumn{1}{c|}{41.12}  & \textbf{ 42.75} \\ 
\multicolumn{1}{c}{\CIRCLE}         & \CIRCLE          & \multicolumn{1}{c|}{49.18}   & \multicolumn{1}{c}{47.23}   & \multicolumn{1}{c}{37.77} & \multicolumn{1}{c}{47.46} & \multicolumn{1}{c}{48.13} & \multicolumn{1}{c|}{48.45}  & \textbf{49.62} \\ \toprule
\multicolumn{2}{c|}{Average}                    & \multicolumn{1}{c|}{44.65}   & \multicolumn{1}{c}{42.77}   & \multicolumn{1}{c}{34.07} & \multicolumn{1}{c}{43.08} & \multicolumn{1}{c}{43.92} & \multicolumn{1}{c|}{44.26}  & \textbf{45.58} \\ \toprule
\end{tabular}
\vspace{-0.5em}
\end{table*}

\begin{table}[]
\centering
\caption{Performance on the multimodal segmentation task with the Cityscapes dataset.}
\label{pe-se-city}
\begin{tabular}{cc|ccc}
\toprule
\multicolumn{2}{c|}{Modality}                   & \multicolumn{3}{c}{mIOU($\uparrow$)}                        \\ \hline
\multicolumn{1}{c}{\multirow{2}{*}{RGB}} & \multirow{2}{*}{Depth} & \multicolumn{1}{c|}{Customized} & \multicolumn{2}{c}{Unified}      \\ \cline{3-5} 
\multicolumn{1}{c}{}           &            & \multicolumn{1}{c|}{ESANet}   & \multicolumn{1}{c|}{mmFormer} & MMANet \\ \toprule
\multicolumn{1}{c}{ \CIRCLE }         &    \Circle         & \multicolumn{1}{c|}{77.60}   & \multicolumn{1}{c|}{76.62}  & \textbf{77.61} \\ 
\Circle            & \CIRCLE           & \multicolumn{1}{c|}{59.11}   & \multicolumn{1}{c|}{58.53}  & \textbf{60.12} \\ 
\multicolumn{1}{c}{ \CIRCLE }         & \CIRCLE           & \multicolumn{1}{c|}{78.62}   & \multicolumn{1}{c|}{78.01}  & \textbf{78.89} \\ \toprule
\multicolumn{2}{c|}{Average}                    & \multicolumn{1}{c|}{71.77}   & \multicolumn{1}{c|}{71.05}  & \textbf{72.20} \\ \toprule
\end{tabular}
\vspace{-0.5em}
\end{table}

\subsection{Performance and Comparison on Multimodal Segmentation}

\textbf{Datasets}: We take the semantic segmentation task as the example of multimodal segmentation and conduct experiments on the NYUv2~\cite{nyuv2} and Cityscapes~\cite{citycape} datasets. Both of them consist of the RGB and Depth modalities. Specifically, NYUv2 contains 1,449 indoor RGB-D images, of which 795 are used for training and 654 for testing. We used the common 40-class label setting. Cityscapes is a large-scale outdoor RGB-D dataset for urban scene understanding. It contains 5,000 finely annotated samples with a resolution of 2048×1024, of which 2,975 for training, 500 for validation, and 1,525 for testing. Cityscapes also provides 20k coarsely annotated images, which \textit{we did not use for training}. For both datasets, we report the results on the validation set using the metric of mean IOU (mIOU).

\textbf{Implementation}: We use random scaling, cropping, color jittering, and flipping for data augmentation. All models are optimized by Adam for 300 epochs with a mini-batch 8. The learning rate is initialized with 1e-2 and adapted by the PyTorch’s one-cycle scheduler~\cite{onecycle}. 


The hyper-parameters of the comparison methods use the suggested ones in their article.The hyper-parameter $(\alpha,\beta)$ for MMANet is set as $(4,0.2)$ and $(10,0.1)$ for the NYUv2 and Cityscapes datasets, respectively. The warm-up epoch $N$ is set as 10.

\textbf{Comparison}: We also compare MMANet with the HeMIS~\cite{hemis}, LCR~\cite{lcr}, RFNet~\cite{rfnet}, and mmFormer~\cite{mmformer}. Here, we set ESANet and ESANnet-MD as the baseline. ESANet~\cite{ESANet} is an efficient and robust model for RGB-D segmentation, which trains the model for each modality combination. ESANet-MD is the variant of ESANet by simply adding the Bernoulli indicator after its modality encoder. ESANet-MD trains only a single model for all modality combinations. Finally, for a fair comparison, we unify the modality encoder of HeMIS, LCR, RFNet, and mmFormer as the ResNet50 with NBt1 used in ESANet. Besides, we set the ESANet model trained with complete multimodal data as the teacher network and the ESANet-MD model as the development network.

\textbf{Results}: Table~\ref{pe-se-nyu} and Table~\ref{pe-se-city} list the comparison results on the NYUv2 and Cityscapes datasets, respectively. From these results, we can see that MMANet achieves the best performance on both datasets for all the modality combinations. In particular, it outperforms the second-best method, mmFormer, by 1.32\% and 1.05\% in the NYUv2 and Cityscapes datasets, respectively. This demonstrates the effectiveness of the MMANet on the multimodal segmentation task. Moreover, MMANet improves the average performance of ESANet-MD by 2.81\% in the NYUv2 dataset and even outperforms the customized baseline, ESANet, by 0.97\% and 0.43\% in NYUv2 and Cityscapes datasets. This shows the effectiveness of the proposed MAD and MAR on the incomplete multimodal segmentation task.


\section{Ablation Study}

This section will study the effectiveness of MAD and MAR and conduct extensive ablation experiments on four datasets. Limited by page, we only present the results of the CASIA-SURF dataset and other results can be seen in the supplementary material.

\subsection{The effectiveness of MAD} 
\begin{figure}[t]
\centering
\includegraphics[width=0.9\columnwidth]{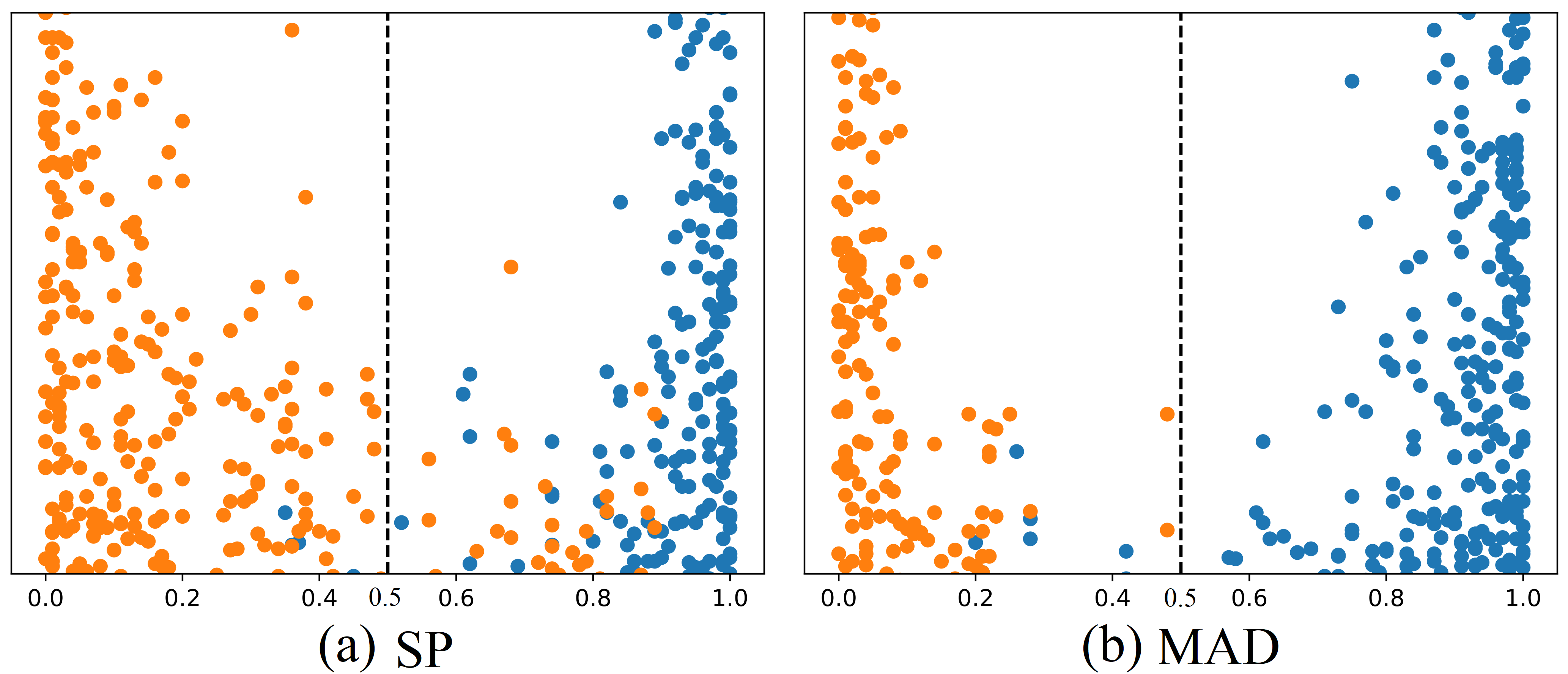} 
\caption{The prediction distribution of the SF-MD model assisted by SP and MAD on CASIA-SURF dataset. X-axis represents the normalized logit output and x=0.5 is the classification boundary. orange and blue dots denotes two different classes.}
\label{MAD_sp_mad}
\vspace{-1.0em}
\end{figure}

To study the effect of MAD, we conduct experiments to compare the performance of the vanilla SF-MD and its variant with SP and MAD. Here, SP is the degradation method of MAD that transfers knowledge by directly matching the cosine distance of the sample representations between teacher and deployment networks. The results are shown in Table~\ref{ab-mad-surf}. We can see that the variant of SF-MD consistently outperforms the vanilla SF-MD in all the modality combination and improve its performance by 1.6\% and 2.93\% in average. This demonstrates the effectiveness of transferring comprehensive multimodal information from the teacher network to the deployment network. Furthermore, the proposed MAD outperforms SP by 1.33\%, which demonstrates the validity of re-weighing sample loss via the classification uncertainty. This is because the classification uncertainty re-weighing can encourage the deployment to focus on the hard samples and thus acquire a more separable inter-class margin than the conventional SP (see Fig.~\ref{MAD_sp_mad}).





\begin{table}[]
\centering
\caption{Ablation result of MAD on the CASIA-SURF dataset.}
\label{ab-mad-surf}
\begin{tabular}{ccc||ccc}
\toprule
RGB   & Depth & IR  & SF-MD & +SP  & +MAD \\ \toprule
\CIRCLE  &\Circle&\Circle& 11.75 & 10.7 & \textbf{10.36} \\ 
    & \CIRCLE &\Circle& 5.87 & 3.3  & \textbf{2.54} \\ 
    &\Circle& \CIRCLE & 16.62 & 15.03 & \textbf{11.67} \\ 
\CIRCLE  & \CIRCLE &\Circle& 4.61 & 2.52 & \textbf{1.23} \\ 
\CIRCLE  &\Circle& \CIRCLE & 6.68 & 5.16 & \textbf{4.09} \\ 
    & \CIRCLE & \CIRCLE & 4.95 & 3.18 & \textbf{1.44} \\
\CIRCLE  & \CIRCLE & \CIRCLE & 2.21 & 1.13 & \textbf{0.77} \\ \toprule
\multicolumn{3}{c||}{Average}& 7.5  & 5.9  & \textbf{4.57} \\ \toprule
\end{tabular}
\end{table}

\subsection{The effectiveness of MAR} To study the effect of MAR, we conduct experiments to compare the performance of the SF-MAD, namely the SF-MD with the MAD, and its variant with SR and MAR. Here SR is the conventional modality regularization strategy considering only the single modality combination. As shown in Table~\ref{ab-mad-surf}, SR and MAR improve the performance of SF-MAD by 0.24\% and 0.63\% in average, respectively, showing the effeteness to regularize the single and weak modality combinations. Moreover, MAR outperforms SR by 0.39\% in average, demonstrating the superiority of MAR.

\begin{figure}[t]
\centering
\includegraphics[width=1.0\columnwidth]{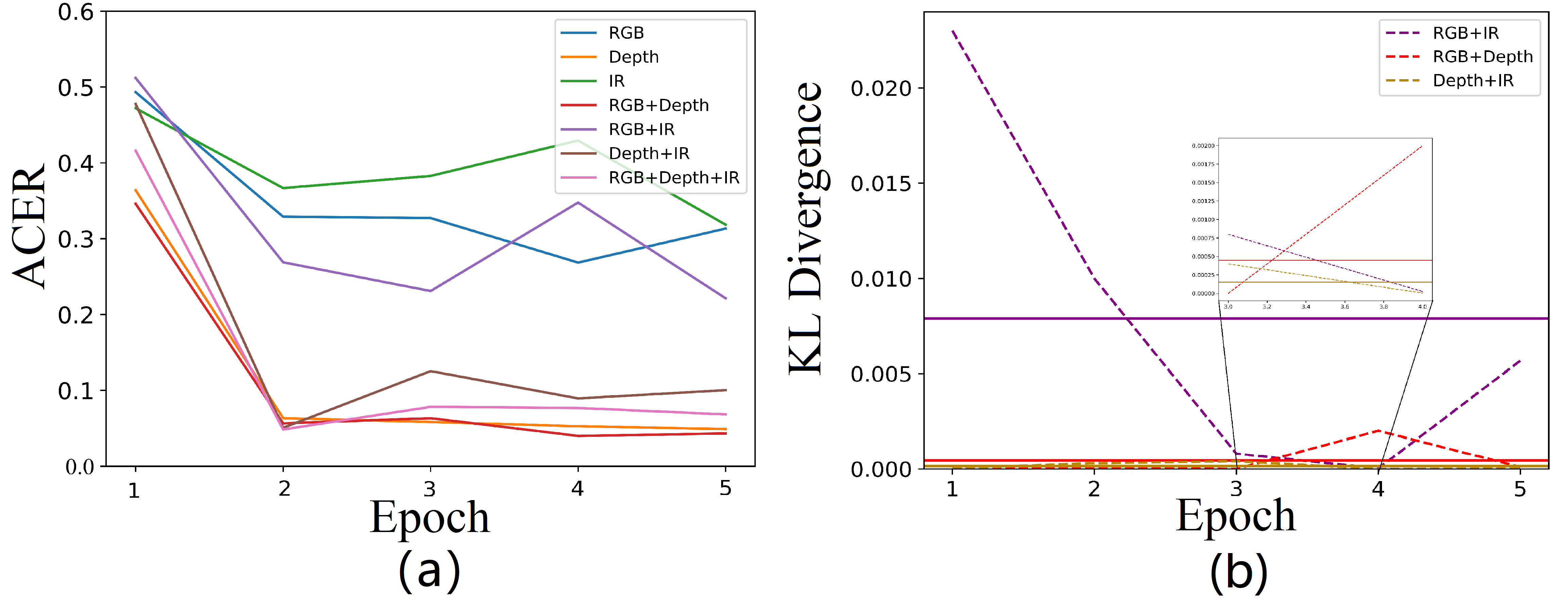} 
\caption{(a) The learning process of different modality combinations on the CASIA-SURF dataset during the warm-up stage. (b) The corresponding $g^{d} \in R^{3}$ for (dashed line) and its average result $\overline{g}^{d}$ (solid line) for the warm-up stage.}
\label{training_pro}
\vspace{-1.0em}
\end{figure}

Here the average gain of SR and MAR is less than SP and MAD since they aim to improve the performance of only the weak, not all combinations. Specifically, as shown in Table~\ref{ab-mad-surf}, the three worst-performing combinations are `RGB', `IR' and, `RGB+IR'. However, SR only focuses on the combinations of single modality, RGB (1.19\%), IR (1.46\%), and Depth (0.65\%), where `Depth' is exactly a strong modality. In contrast, Fig.~\ref{training_pro}(b) shows that the prediction discrepancy between `RGB+IR' and `RGB+Depth+IR' is the largest. And the performance gain of MAR mainly comes from RGB (1.79\%), IR (1.63\%), as well as the combination of RGB and IR (1.02\%). These results show that MAR can mine the weak modality combinations more accurately and force the deployment network to improve its discrimination ability for them.



\begin{table}[]
\centering
\caption{Ablation result of MAR on the CASIA-SURF dataset.}
\label{ab-maR-surf}
\begin{tabular}{ccc||ccc}
\toprule
RGB   & Depth & IR  & SF-MAD & +SR  & +MAR \\ \toprule
\CIRCLE  &\Circle&\Circle& 10.36 & 9.17 & \textbf{8.57} \\ \hline
    & \CIRCLE &\Circle& 2.54  & \textbf{1.89} & 2.27 \\
    &\Circle& \CIRCLE & 11.67 & 10.21 & \textbf{10.04} \\ \hline
\CIRCLE  & \CIRCLE &\Circle& \textbf{1.23}  & 1.66 & 1.61 \\ 
\CIRCLE  &\Circle& \CIRCLE & 4.09  & 4.37 & \textbf{3.01} \\ \hline
    & \CIRCLE & \CIRCLE & 1.44  & 2.12 & \textbf{1.18} \\
\CIRCLE  & \CIRCLE & \CIRCLE & \textbf{0.77}  & 0.92 & 0.87 \\ \toprule
\multicolumn{3}{c||}{Average}& 4.57  & 4.33 & \textbf{3.94} \\ \toprule
\end{tabular}
\end{table}

\section{Conclusion}


This paper presents an MMANet framework to aid the deployment network for incomplete multimodal learning. Specifically, MMANet introduces a teacher network pre-trained with complete multimodal data to transfer the comprehensive multimodal information to the deployment network via MAD. This helps it acquire modality-invariant and specific information while maintaining robustness for incomplete modality input. Besides, MMANet introduces a regularization network to mine and regularize weak modality combinations via MAR. This forces the deployment network to improve its discrimination ability for them effectively and adaptively. Finally, extensive experiments demonstrate the effectiveness of the proposed MMANet, MAD, and MAR for incomplete multimodal learning.




{\small
\bibliographystyle{ieee_fullname}
\bibliography{egbib}
}

\end{document}